# An Automatic System to Monitor the Physical Distance and Face Mask Wearing of Construction Workers in COVID-19 Pandemic

[1]Moein Razavi*
Department of Computer Science and Engineering
Texas A&M University
moeinrazavi@tamu.edu

Hamed Alikhani*
Department of Engineering
Texas A&M University
hamedalikhani@tamu.edu

Vahid Janfaza*
Department of Computer Science and Engineering
Texas A&M University
vahidjanfaza@tamu.edu

Benyamin Sadeghi
Industrial and Manufacturing Engineering
University of Wisconsin-Milwaukee
bsadeghi@uwm.edu

Ehsan Alikhani
Systems Science and Industrial Engineering
State University of New York at Binghamton
ealikha1@binghamton.edu

**Abstract**

*The COVID-19 pandemic has caused many shutdowns in different industries around the world. Sectors such as infrastructure construction and maintenance projects have not been suspended due to their significant effect on people's routine life. In such projects, workers work close together that makes a high risk of infection. The World Health Organization recommends wearing a face mask and practicing physical distancing to mitigate the virus's spread. In this paper, we developed a computer vision system to automatically detect the violation of face mask wearing and physical distancing among construction workers to assure their safety on infrastructure projects during the pandemic. For the face mask detection, we collected and annotated 1,000 images, including different types of face mask wearing, and added them to a pre-existing face mask dataset to develop a dataset of 1,853 images and increased the dataset to 3300 images by data augmentation. Then we trained and tested multiple Tensorflow state-of-the-art object detection models on the face mask dataset and chose the Faster R-CNN Inception ResNet V2 network that yielded the accuracy of 99.8%. For physical distance detection, we employed the Faster R-CNN Inception V2 to detect people. A transformation matrix was used to eliminate the camera angle's effect on the object distances on the image. The Euclidian distance used the pixels of the transformed image to compute the actual distance between people. A threshold of six feet was considered to capture physical distance violation. We also used transfer learning for training the model. The final model was applied on four videos of road maintenance projects in Houston, TX, that effectively detected the face mask and physical distance. We recommend that construction owners use the proposed system to enhance construction workers' safety in the pandemic situation.*

## 1. Introduction

The spread of COVID-19 has resulted in more than 1,841,000 global deaths and more than 351,000 deaths in the US by Dec. 31, 2020 [1]. The spread of virus can be avoided by mitigating the effect of the virus in the environment [2], [3] or preventing the virus transfer from person to person by practicing physical distance and wearing face masks. WHO defined physical distancing as keeping at least six feet or two meters distance from others and recommended that keeping the physical distance and wearing a face mask can significantly reduce transmission of the COVID-19 virus [4]–[7]. Like other sectors, the construction industry has been affected, where unnecessary projects have been suspended or mitigated people's interaction. However, many infrastructure projects cannot be suspended due to their crucial role in people's life. Therefore, bridge maintenance, street widening, highway rehabilitation, and other essential infrastructure projects have been activated again to keep the transportation system's serviceability. Although infrastructure projects are activated, the safety of construction workers cannot be overlooked. Due to the high density of workers in construction projects, there is a high risk of the infection spread in construction sites [8]. Therefore, systematic safety monitoring in infrastructure projects that ensure maintaining the physical distance and wearing face masks can enhance construction workers' safety.

Safety agents are sometimes deployed to infrastructure projects to inspect workers to see whether they are complying with social distancing or wearing face masks. However, once there are so many workers on a construction site, it is difficult for the officers to determine hazardous situations. Also, assigning safety officers increases the number of people on-site, raising the chance of transmission even more, and putting workers and officers in a more dangerous situation. Recently, online video capturing in construction sites has become very common.

---

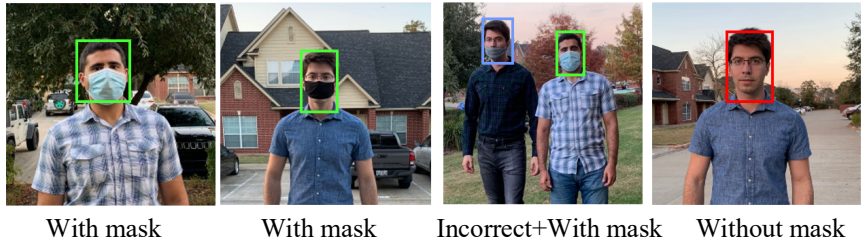

With mask     With mask     Incorrect+With mask     Without mask

Figure 1. Examples of images in the facemask database

Drones are used in construction projects to record online videos to manage worksites more efficiently [9]–[11]. The current system of online video capturing can be used for safety purposes. An automatic system that uses computer vision techniques to capture real-time safety violations from online videos can enhance infrastructure project workers' safety. This study develops a model using Faster R-CNN to detect workers who either do not wear a face mask or do not maintain the physical distance in road projects. Once a safety violation occurs, the model highlights who violates the safety rules by a red box in the video.

## 2. Literature Review

Object detection problems aim to locate and classify objects in an image [12]. The face mask and physical distance detection are classified as object detection problems. Object detection algorithms have been developing in the past two decades. Since 2014, deep learning usage in object detection has driven remarkable breakthroughs, improving accuracy and detection speed [13].

Object detection is divided into two categories. One-stage detection, such as You Only Look Once (YOLO), and two-stage detection, such as Region-Based Convolutional Neural Networks (R-CNN). Two-stage detectors have higher localization and object recognition accuracy, while the one-stage detectors achieve greater inference speed [14]. R-CNN or regions with CNN features (R-CNNs), introduced by Girshick et al. [15], implements four steps; First, it selects several regions from an image as object candidate boxes, then rescales them to a fixed size image. Second, it uses CNN for feature extraction of each region. Finally, the features of each region are used to predict the category of boundary boxes using the SVM classifier [15], [16]. However, feature extraction of each region is resource-intensive and time-consuming since candidate boxes have overlap, making the model perform repetitive computation. Fast R-CNN handles the issue by taking the entire image as the input to CNN to extract features [17]. Faster R-CNN speeds up the R-CNN network by replacing selective search with a Region Proposal Network (RPF) to reduce the number of candidate boxes [18]–[20]. Faster R-CNN is a near-real-time detector [18].

Face mask detection identifies whether a person is wearing a mask or not in a picture [21], [22]. Physical distance detection first recognizes people in a picture then identifies the real distance between them [23]. Since the beginning of COVID-19 pandemic, studies have been conducted to detect face masks and physical distancing in the crowd. Jiang et al. [21] employed a one-stage detector, called RetinaFaceMask, that used a feature pyramid network to fuse the high-level semantic information. They added an attention layer to detect the face mask faster in an image. They achieved a higher accuracy of detection comparing with previously developed models. Militante & Dionisio [24] used the VGG-16 CNN model and achieved 96% of accuracy to detect people who wear a face mask or not. Rezaei & Azarmi [25] developed a model based on YOLOv4 to detect face mask wearing and physical distancing. They trained their model on some accessible big databases and achieved the precision of 99.8% for a real-time detection. Ahmed et al. [23] employed YOLOv3 to detect people in the crowd then used the Euclidian distance to detect the physical distance between two people. They also used a tracking algorithm to track people who violate the physical distancing in a video. They achieved an initial accuracy of 92% and then increased the accuracy by applying transfer learning to 98%.

One major problem in training of deep learning models for construction sites is data deficiency. Using drones in construction sites is one of the methods to address this issue. In construction projects, drones capture real-time video and provide aerial insights that help catch problems immediately. Research studies have collected large scale image data from drones and applied deep learning techniques to detect objects. Asadi & Han [26] developed a system to improve data acquisition from drones in construction. Mirsalar Kamari & Ham [10] used drones' image data in a construction site to quickly detect areas vulnerable to the wind in the job site. Li et al. [27] applied deep learning object detection networks on video captured from drones to track moving objects. The data obtained from drones can be used to detect workers and identify whether they are wearing face masks and practice physical distancing. In this paper, to address the data deficiency problem, we used different data augmentation techniques.

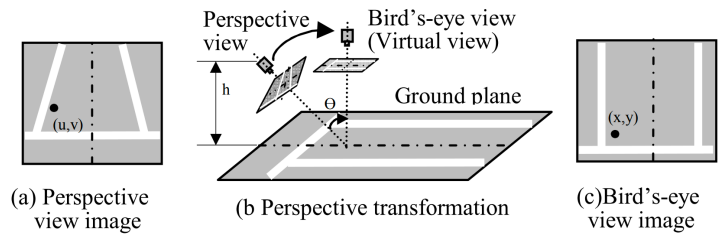

(a) Perspective view image     (b) Perspective transformation     (c) Bird's-eye view image

Figure 2. A perspective image transformation to a bird's

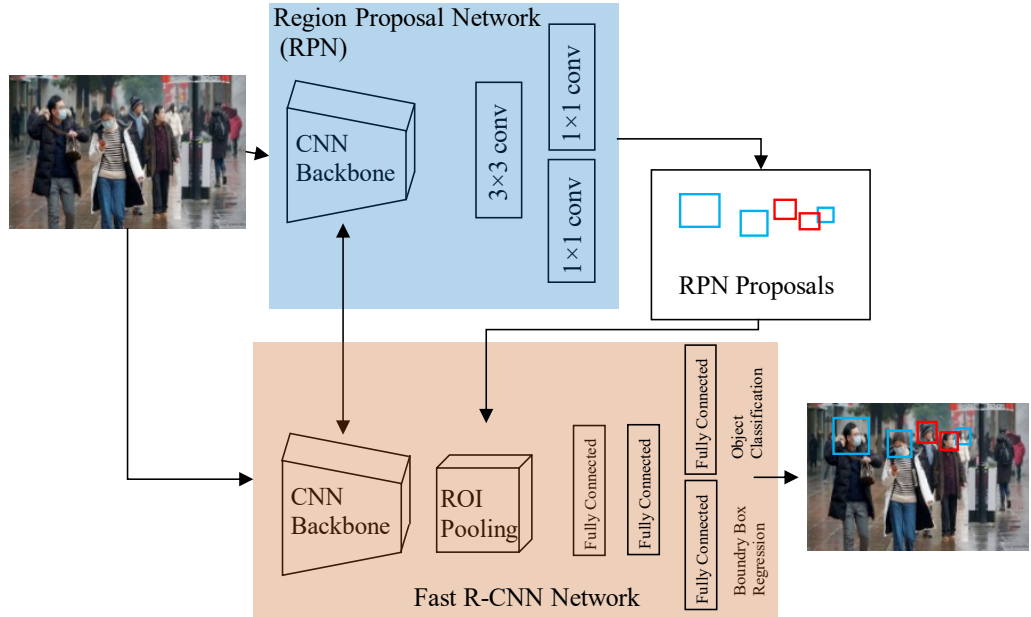

Figure 3. A schematic architecture of the Faster R-CNN

## 3. Methodology

We obtained a facemask dataset available on the internet and increased the size of dataset by adding more images and applying data augmentation. Then, we trained multiple Faster R-CNN object detection models to choose the most accurate model for face mask detection. For the physical distance detection, we used a Faster R-CNN model to detect people and then used the Euclidian distance to obtain the people's distance in reality based on the pixel numbers in the image. Then, we exploited transfer learning to increase the accuracy. The model was applied on multiple videos of road maintenance projects in Houston, TX, to validate the performance of the model.

### 3.1. Dataset

A part of the dataset of face masks was obtained from MakeML website [28] that contains 853 images that each image includes one or multiple normal faces with various illumination and poses. The images are already annotated with faces with a mask, without mask, and incorrect mask wearing. To increase the training data 1,000 other images with their annotations were added to the database. The total of 1,853 images was used as the facemask dataset (#with mask: 1013, #without mask: 717, #mask worn incorrect: 123). To avoid class imbalance and to increase the size of training dataset, we applied a combination of rotation, flipping, contrast, and brightness data augmentation techniques. This would also help with the avoidance of overfitting and making the model robust in the detection of new unseen data. After data augmentation, we increased the number of images in each class to 1100 (3300 total). Some samples of images with their annotations are illustrated in Figure 1, where three types of face mask wearing are annotated including a correct face mask wearing, incorrect wearing, and without face mask.

### 3.2. Face mask detection

For the face mask detection task, five different object detection models in the Tensorflow object detection model Zoo [29] were trained and tested on the face mask dataset to compare their accuracy and choose the best model for the face mask detection. The dataset was split into 80% and 20% for training and testing. To calculate the accuracy, we considered the prediction correct if the model predicts the true label with a confidence of at least 80%. Table 1 shows the models and their associated accuracies.

Table 1. Object detection models and their accuracy on the Face Mask dataset

| # | Model | Image size | Accuracy |
|---|---|---|---|
| 1 | Faster R-CNN Inception ResNet V2 | 800*1333 | 99.8% |
| 2 | Faster R-CNN Inception ResNet V2 | 640*640 | 81.8% |
| 3 | Faster R-CNN ResNet 152 V1 | 640*640 | 95% |
| 4 | SSD ResNet50 V1 FPN (RetinaNet50) | 640*640 | 82% |
| 5 | SSD MobileNet V1 FPN | 640*640 | 93.6% |

The Faster R-CNN Inception ResNet V2 800*1333 was selected due to its highest accuracy, i.e., 99.8%.

### 3.3. Physical distance detection

We used the physical distancing detector model introduced by Roth [31]. The model detects the physical distancing in three steps; people detection, picture transformation, and distance measurement. Roth [32] trained models available on the Tensorflow object detection model Zoo on the COCO data set that includes 120,000 images. Among all the models, the Faster R-CNN Inception V2 with COCO weights was selected as people detection through model evaluation due to its highest detector performance indicator.

The proposed method in this paper is applicable for a fixed camera. To detect the 6-ft. distance, in the first-time use, the application will ask the user to indicate the 4 corners of a square with edge-length = 6-ft. on the ground using mouse input. Then, using a transformation matrix, the application will transform the perspective view to bird's eye view which can be applied to any point on the ground. Therefore, since the resolution and the number of pixels in the image are known, the application can calculate the number of pixels for a 6-ft. distance in an arbitrary image. Figure 2 shows the original image captured from a perspective to the vertical view of bird's eye, where the dimensions in the picture have a linear relationship with real dimensions [30]. The relationship between pixel of (x, y) in the bird's eye picture and pixel of (u, v) in the original picture is defined as:

$$\begin{bmatrix} x' \\ y' \\ w' \end{bmatrix} = \begin{bmatrix} a_{11} & a_{12} & a_{13} \\ a_{21} & a_{22} & a_{23} \\ a_{31} & a_{32} & a_{33} \end{bmatrix} \begin{bmatrix} u \\ v \\ w \end{bmatrix}$$

Where $x = x'/w'$ and $y = y'/w'$. For the transformation matrix, the OpenCV library in Python was used [33]. Finally, the distance between each pair of people is measured by estimating the distance between the bottom-center point of each boundary box in the bird's eye view.

### 3.4. Faster R-CNN model

Figure 3 shows a schematic architecture of the Faster R-CNN model. The Faster R-CNN includes the Region Proposal Network (RPN) and the Fast R-CNN as the detector network. The input image is passed through the Convolutional Neural Networks (CNN) Backbone to extract the features. The RPN then suggests bounding boxes that are used in the Region of Interest (ROI) pooling layer to perform pooling on the image features. Then, the network passes the output of the ROI pooling layer through two Fully Connected (FC) layers. The first FC layer determines the class of each object and the second one performs a regression to improve the proposed boundary boxes [17], [18], [34].

### 3.5. Transfer learning

The size of the training dataset for the face mask detection is limited. Since the face mask detection model is a complex network, the scarce data would result in an overfitted model. Transfer learning is a common technique in machine learning when the dataset is limited and the training process is computationally expensive. Transfer learning uses the weights of a pre-trained model on a big dataset in a similar network as the starting point in training [35]. In this research, we used the weights of the Tensorflow pre-trained object detection models with for the model training. These models have been pretrained on COCO dataset and COCO detection metrics were used for evaluation.

### 4. Results and discussion

For both networks of face mask detection and physical distance recognition the Google Colaboratory was used. Google Colaboratory is a cloud service developed by Google Research that provides python programming environment for executing machine learning and data analysis codes and provides free access to different types of GPUs including Nvidia K80s, T4s, P4s and P100s and includes leading deep learning libraries [36]. In this

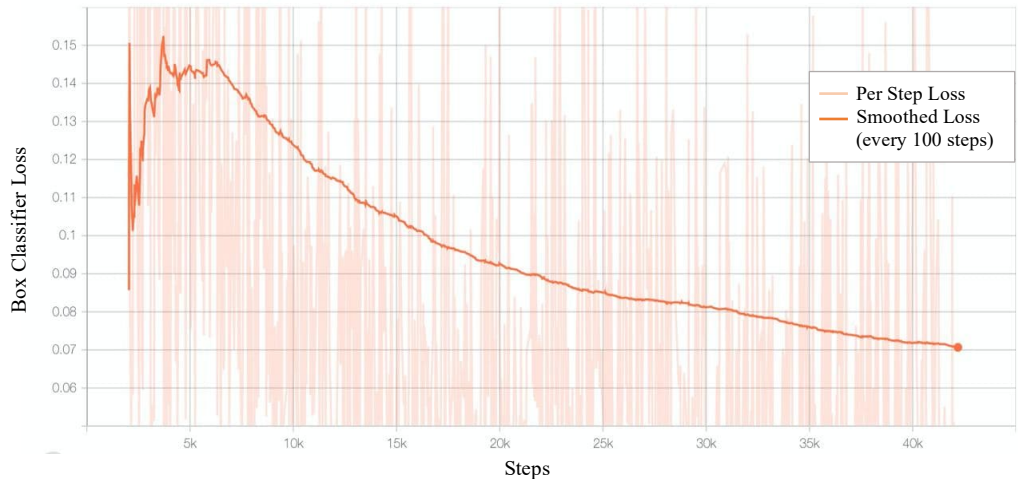
Figure 4. The convergence of the classification loss for the face mask detection model

experiment, for the face detection, we used the batch size of 1, the momentum optimizer of value 0.9 with cosine decay learning rate (learning rate base of 0.008), and the image size of 800*1333. The maximum number of steps was 200,000 and the training of the model was stopped when the classification loss reached below 0.07, that happened in near the 42,000$^{th}$ step (Figure 4). For the physical distance detection model, we used the batch size of 1, the total number of steps of 200,000, and the momentum optimizer of value 0.9 with manual step learning rate. The first step was from zero to 90,000, where the learning rate was 2e-4, the second step was from 90,000 to 120,000, where the learning rate was 2e-5, and the third step was from 120,000 to 200,000, where the learning rate was 2e-6.

The two models of face mask detection and physical distance recognition were combined. Four videos of road maintenance projects captured in Houston, TX, were given to the model to test the performance of the model on construction workers. Figure 5 shows screenshots of one moment of each case. In case 1, the model detected a worker without a mask with 99% confidence and a worker wearing a mask with 97% confidence. The model also highlighted workers who do not practice physical distancing with a red box indicating a zone with a high risk of infection and a person far from others with a green box suggesting a safe zone. In case 2, the model detected an incorrect mask wearing and a correct mask wearing with 89% and 94% confidence, respectively. The model indicated that the workers are keeping the physical distance. The slight drop in the confidence of the incorrect mask wearing is because of the smaller number of training data for the incorrect face mask wearing case in the dataset. Cases 3 and 4 indicate high confidence in detecting the face masks and high accuracy in physical distance detection. In case 3, the model detected a face mask for the worker on the left side of the image with a relatively lower confidence due to the lower number of training data for a rotating head with this type of mask wearing. However, the output cases indicate reliable confidence in detection of face masks and accuracy of the physical distance detection for the workers in road projects. For the cases that the face is not visible (e.g., Case #4), the model will only consider the social distancing criterion. In all four cases, the processing time of the face mask and physical distancing detection took lower than the length of each video. It suggests that the model is able to work in real-time that is a handful feature for safety officers to instantly detect any violation.

## 5. Conclusion

In this paper, we developed a model to detect face mask wearing and physical distancing among construction workers to assure their safety in the COVID-19 pandemic. We collected a facemask dataset including images of people with mask, without mask, and incorrect mask wearing. To increase the training dataset, 1,000 images with different types of mask wearing were collected and added to the dataset. Then, we applied a combination of different data augmentation methods and increased the size of the dataset to 3300 images in total. A Faster R-CNN Inception ResNet V2 network was chosen among different models that yielded the accuracy of 99.8% for face mask detection. For physical distance detection, the Faster R-CNN Inception V2 was used to detect people and a transformation matrix was used to remove the effect of the camera angle on actual distance. The Euclidian distance converted the pixels of the transformed image to people's real distance. The model set a threshold of six feet to capture physical distance violation. Transfer learning was used for training models. Four videos of actual road maintenance projects in Houston, TX, were used to evaluate the combined model. The output of the four cases indicated an average of more than 90% accuracy in detecting different types of mask wearing in construction workers. Also, the model accurately detected workers who were too close and did not practice the physical distancing. Road project owners and contractors can use the model results to monitor workers to avoid infection and enhance

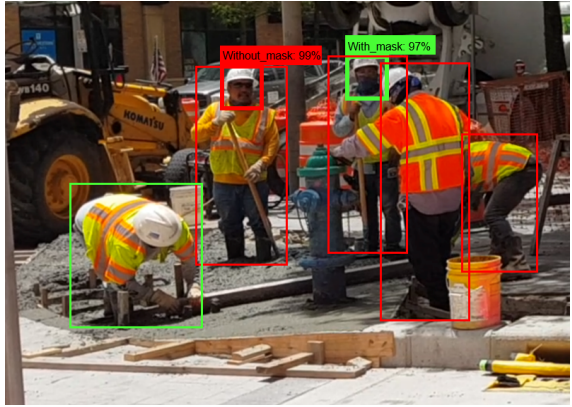 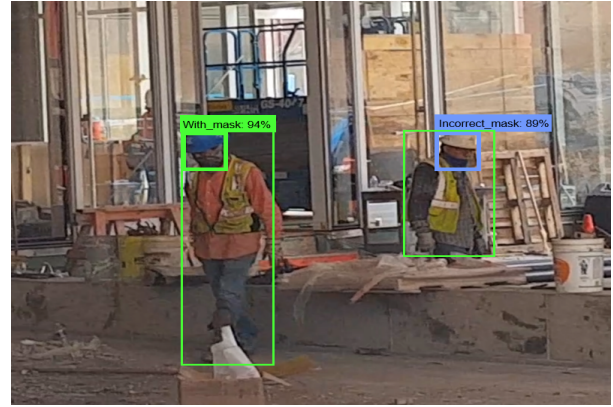

Case #1        Case #2

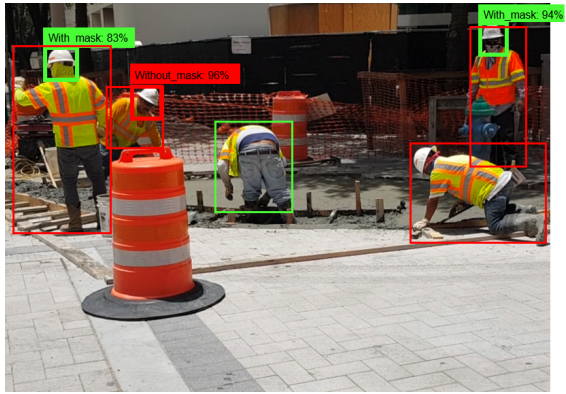 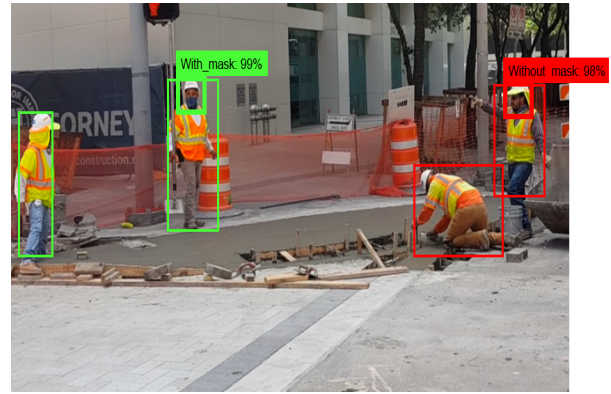

Case #3        Case #4

Figure 5. The application of the model on four road construction cases

workers' safety. Future studies can employ the model on other construction projects such as building projects. Also, future studies can try other detection models and tune the hyper-parameters to increase the detection accuracy.

Compliance with Ethical Standards:
There was no funding was dedicated to this project.

Conflict of Interest:
There was no conflict of interest related to this project.

Ethical approval: This article does not contain any studies with human participants performed by any of the authors.